\documentclass{article}

\usepackage[preprint]{neurips_2026}

\PassOptionsToPackage{numbers, compress}{natbib}

\usepackage[utf8]{inputenc} 
\usepackage[T1]{fontenc}    
\usepackage{hyperref}       
\usepackage{url}            
\usepackage{booktabs}       
\usepackage{amsfonts}       
\usepackage{nicefrac}       
\usepackage{microtype}      
\usepackage{xcolor}         
\usepackage{pifont}
\newcommand{\cmark}{\ding{51}}
\newcommand{\xmark}{\ding{55}}
\usepackage{tikz}
\def\halfcheckmark{\tikz\draw[scale=0.35,fill=black](0,.35) -- (.25,0) -- (1,.7) -- (.25,.15) -- cycle (0.75,0.2) -- (0.77,0.2)  -- (0.6,0.7) -- cycle;}
\newcommand{\rao}[1]{\renewcommand{\arraystretch}{#1}}

\usepackage{multirow}
\usepackage{xurl}   
\usepackage{pifont} 
\usepackage{subcaption}
\usepackage{wrapfig}
\usepackage{enumitem}
\usepackage{amsmath,amssymb}
\usepackage{algorithmic}
\usepackage{graphicx}
\usepackage{mathtools}
\usepackage{multicol}
\usepackage{tabularx}

\title{LLMSpace: Carbon Footprint Modeling for Large Language Model Inference on LEO Satellites}

\author{
\begin{tabular}{cccc}
Lei Jiang & Adrian Ildefonso & Daniel Loveless & Fan Chen\\
\end{tabular}\\
Indiana University \\
\texttt{\{jiang60,aildefon,dlovele,fc7\}@iu.edu}
}

\begin{document}

\maketitle

\begin{abstract}
Large language models (LLMs) impose rapidly growing energy demands, creating an emerging energy and carbon crisis driven by large-scale inference. Solar-powered, AI-enabled low Earth orbit (LEO) satellites have been proposed to mitigate terrestrial electricity consumption, but their lifecycle carbon footprint remains poorly understood due to launch emissions, satellite manufacturing, and radiation-hardened hardware requirements. This paper presents \textit{LLMSpace}, the first carbon modeling framework for LLM inference on AI-enabled LEO satellites. LLMSpace jointly models operational and embodied carbon, peripheral subsystems, radiation-hardened accelerators and memories, and LLM-specific workload characteristics such as prefill–decode behavior and token generation. Using realistic satellite and GPU configurations, LLMSpace reveals key trade-offs among carbon footprint, inference latency, hardware design, and operational lifetime for sustainable space-based LLM inference. Source code: \url{https://github.com/UnchartedRLab/LLMSpace}.
\end{abstract}

\section{Introduction}
\vspace{-0.1in}

LLMs are among the most energy-intensive AI systems due to their large-scale deployment, raising significant environmental concerns. A single short GPT-4o query consumes $\sim0.43$~Wh, corresponding to an estimated 392--463~GWh of annual electricity consumption for GPT-4o inference in 2025~\cite{jegham2025hungry}—comparable to the yearly usage of 35K average U.S.\ households. As LLMs become increasingly embedded in everyday applications, including web browsers and operating systems, their ubiquity continues to grow. Consequently, global data center electricity demand is projected to more than double by 2030~\cite{pilz2025trends}, further amplifying AI-related carbon emissions.

To address these challenges, major cloud providers—including Microsoft~\cite{microsoft_azure_space_2025}, StarCloud~\cite{lee_starcloud_2025}, Google~\cite{beals_suncatcher_2025}, SpaceX, and Amazon~\cite{amazon_leo_website_2025}—have proposed solar-powered low Earth orbit (LEO) satellite constellations for communication and connectivity. Building on this infrastructure, recent academic and industry efforts~\cite{wang2024case,Bleier:MICRO2023,aili2025development} explore AI-enabled satellites equipped with accelerators (e.g., GPUs~\cite{slater2020total} or TPUs~\cite{casey2023single}) to offload LLM inference to space, leveraging continuous solar energy to reduce reliance on terrestrial power grids. Proposed LEO systems span a wide range of scales, from compact CubeSats~\cite{liddle2020space} (volumes $\sim$$0.001~\mathrm{m}^3$, masses of a few kilograms) to large platforms such as Starlink satellites~\cite{chaudhry2021laser} (hundreds of kilograms). Typically deployed at altitudes of several hundred to $\sim$2{,}000~km via launch vehicles (e.g., Falcon-9~\cite{ohs2025dirty}), these satellites employ heterogeneous communication interfaces~\cite{chaudhry2021laser} for ground and inter-satellite links, operating individually or within constellations to deliver scalable, continuous LLM inference services.

Although solar power can substantially reduce the operational carbon emissions of space-based LLM inference, the embodied carbon arising from launch operations, satellite manufacturing, and radiation-hardened hardware fabrication may partially or fully offset these benefits~\cite{aili2025development}. Despite this trade-off, there is currently no comprehensive carbon modeling framework for AI-enabled LEO satellites that jointly captures operational emissions, embodied carbon, and inference latency. While prior studies have extensively quantified the carbon footprint of LLM training~\cite{llmcarbon} and inference~\cite{fu2025llmco2,jegham2025hungry} in terrestrial data centers, only limited work~\cite{ohs2025dirty,aili2025development} considers space-based computing, and it exhibits several key limitations:
\begin{itemize}[leftmargin=*, nosep, topsep=0pt, partopsep=0pt]
\item \textbf{Incomplete peripheral modeling.} Existing space-based carbon models either omit the embodied carbon of critical peripherals~\cite{ohs2025dirty} (e.g., cooling panels) or provide only coarse estimates by aggregating multiple peripherals into a single value~\cite{aili2025development}. Moreover, prior studies consider a limited set of peripheral configurations~\cite{aili2025development,ohs2025dirty} and overlook alternative designs, such as GaAs and multi-junction solar arrays and radiative cooling panels with integrated heat pipes.

\item \textbf{No support for radiation-hardened computing hardware.} 
Prior models~\cite{ohs2025dirty,aili2025development} assume commercial off-the-shelf (COTS) CMOS devices and neglect space-specific reliability constraints, including total ionizing dose (TID)~\cite{slater2020total}, displacement damage (DD)~\cite{srour2013displacement}, and single-event effects (SEE)~\cite{yue2015modeling}. In orbit, COTS GPUs~\cite{slater2020total} typically operate for only $\sim2$ years, while COTS DRAM~\cite{lee2023investigation} may fail within months. Sustained LLM inference therefore requires radiation-hardened hardware, whose embodied carbon is not captured by existing models.

\item \textbf{Absence of LLM-specific workload modeling.} Existing satellite carbon frameworks~\cite{ohs2025dirty,aili2025development} target conventional workloads (e.g., image processing~\cite{Bleier:MICRO2023}) and do not model the execution characteristics of LLM inference. They do not distinguish between prefill and decoding phases, nor do they incorporate workload-dependent parameters such as prompt length and generated token count. Consequently, they cannot accurately estimate the carbon footprint of diverse LLM applications—including chat, code generation, and summarization—deployed in LEO environments.
\end{itemize}


This paper presents \textit{LLMSpace}, a carbon footprint modeling framework for LLM inference on AI-enabled LEO satellites. Our contributions are summarized as follows:
\begin{itemize}[leftmargin=*, nosep, topsep=0pt, partopsep=0pt]
\item \textbf{Comprehensive peripheral modeling.} \textit{LLMSpace} quantifies the embodied carbon of key subsystems, including solar arrays, batteries, and radiative cooling panels, and supports multiple peripheral configurations across satellite classes with different operational lifetimes.

\item \textbf{Radiation-hardened accelerator and memory modeling.} \textit{LLMSpace} models the embodied carbon of AI accelerators fabricated using radiation-hardened process technologies~\cite{chatzikyriakou2017total}, such as fully depleted silicon-on-insulator (FD-SOI). It also accounts for embodied carbon overheads introduced by architectural mitigation techniques for DD and SEE~\cite{jeon2010evaluation}, including error-correcting codes and triple modular redundancy. In addition, the framework captures radiation-hardened HBM designs for LLM inference based on MRAM technologies~\cite{marinella2021radiation}.

\item \textbf{LLM inference-aware carbon analysis.} \textit{LLMSpace} differentiates the amortized embodied carbon contributions of the prefill and decoding phases of LLM inference. It further models the carbon footprint of diverse LLM workloads—including conversational AI, code generation, and summarization—and identifies workloads that are carbon-efficient in LEO environments.
\end{itemize}
Validations show that \textit{LLMSpace} greatly improves embodied carbon estimation accuracy over prior work. We further demonstrate use cases that analyze trade-offs among embodied carbon, operational energy, and inference latency for LEO-based LLM inference.

\vspace{-0.1in}
\section{Background}
\vspace{-0.1in}

\begin{figure}[t!]
\centering
\includegraphics[width=0.95\textwidth]{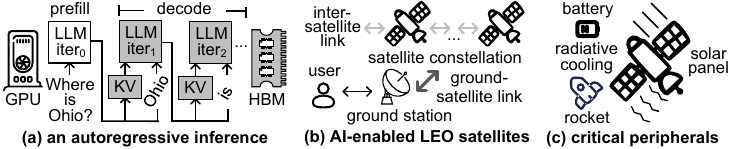}
\vspace{-0.05in}
\caption{AI-enabled LEO satellite platforms for large-scale LLM inference.}
\label{f:carbon_leo_overview}
\vspace{-0.2in}
\end{figure}

\textbf{LLM Inference.} As illustrated in Figure~\ref{f:carbon_leo_overview}(a), LLM inference generates output tokens autoregressively. In the initial iteration, all prompt tokens are processed in parallel to produce the first output token, referred to as the \emph{prefill} phase, during which attention contexts are computed and stored in the key--value (KV) cache. Subsequent iterations form the \emph{decode} phase, where each new token is generated based on the most recent token and cached KV states. The prefill phase is compute-bound and dominated by accelerator throughput, whereas the decode phase is memory-bound and constrained by HBM accesses. Overall energy consumption is substantial and typically dominated by decoding~\cite{wang2025systematic}. For instance, a short GPT-4o request consumes approximately 0.43~Wh, corresponding to a projected annual electricity demand of 392--463~GWh in 2025~\cite{jegham2025hungry}, thereby imposing significant environmental burdens and increasing stress on power infrastructures.

\textbf{AI-Enabled Satellites for LLM Inference.} To alleviate the energy demands of large-scale LLM deployment, recent studies~\cite{wang2024case,Bleier:MICRO2023,aili2025development} have proposed solar-powered, AI-enabled satellites in LEO. These platforms integrate GPUs or TPUs and are deployed via multi-satellite launch vehicles. Owing to scale and radiation constraints, onboard LLM training remains impractical; consequently, satellites primarily support inference workloads~\cite{wang2024case}. As shown in Figure~\ref{f:carbon_leo_overview}(b), user requests are transmitted to ground stations and relayed through ground--satellite links, while satellites may operate independently or coordinate within constellations via inter-satellite links. By relying exclusively on solar energy~\cite{Bleier:MICRO2023}, AI-enabled LEO satellites have the potential to reduce terrestrial electricity demand associated with large-scale LLM inference.

\textbf{Embodied Carbon of AI-Enabled LEO Satellites.} The carbon footprint of LLM inference on LEO satellites consists of operational and embodied components. Operational emissions arise from energy consumption during hardware execution~\cite{fu2025llmco2}. Because AI-enabled LEO satellites are powered exclusively by solar energy, in-orbit operational emissions are effectively negligible. Embodied emissions, in contrast, originate from launch and re-entry, satellite platform manufacturing, peripheral fabrication, and the production of computing and communication hardware. Compared with terrestrial data centers powered entirely by renewable energy, AI-enabled LEO satellites exhibit substantially higher embodied carbon footprints~\cite{Bleier:MICRO2023,aili2025development}. As illustrated in Figure~\ref{f:carbon_leo_overview}(c), beyond launch into LEO, space-based inference systems require additional peripherals—including solar arrays for power generation, batteries for energy storage, and radiative cooling panels for thermal regulation—each contributing to embodied emissions~\cite{aili2025development}.

\textbf{Hardware Reliability in Space.} CMOS circuits in space face reliability challenges due to prolonged exposure to ionizing radiation and energetic particles~\cite{slater2020total}. The primary failure mechanisms include total ionizing dose (TID)~\cite{slater2020total}, displacement damage (DD)~\cite{srour2013displacement}, and single-event effects (SEE)~\cite{yue2015modeling}. TID causes cumulative degradation via charge trapping, leading to threshold shifts, leakage increase, and eventual failure in circuits. DD results from lattice defects induced by non-ionizing energy loss, degrading carrier lifetime and device performance. SEEs arise from high-energy particle strikes, causing transient faults, data corruption, and potential latch-up or burnout. These effects collectively limit AI accelerators' performance and mission lifetime, necessitating radiation-hardened designs.

\textbf{Radiation Hardening Techniques.} 
Mitigating TID, DD, and SEE requires both process- and architectural-level hardening~\cite{kannaujiya2024radiation}. Process-level techniques, such as edgeless CMOS layouts and FD-SOI technologies~\cite{chatzikyriakou2017total}, reduce charge collection and improve radiation tolerance, with additional material engineering enhancing resistance to DD~\cite{Srour:TNS2003}. Due to fabrication complexity, these approaches are typically implemented at mature nodes (e.g., 14~nm)~\cite{boss14nm}, resulting in larger die areas and higher embodied carbon. Architectural techniques—including error-correcting codes (ECCs), triple modular redundancy (TMR), and latch-up protection~\cite{jeon2010evaluation}—further improve resilience but introduce redundancy and area overhead, increasing the embodied carbon footprint.

\vspace{-0.1in}
\section{Related Work and Motivation}
\vspace{-0.1in}

\begin{wraptable}[8]{r}{0.5\linewidth}
\vspace{-0.2in}
\caption{Comparing \textit{LLMSpace} against prior work.} 
\label{t:related_work_comparison}
\vspace{-0.1in}
\rao{0.9}
\setlength{\tabcolsep}{2pt}
\footnotesize 
\begin{center}
\begin{tabular}{ccccc} \toprule
\multirow{2}{*}{scheme}                         & space-    & peripheral     & rad-hard  & LLM-    \\
                                                & based     & breakdown      & hardware  & specific\\ \midrule
\cite{llmcarbon,fu2025llmco2,jegham2025hungry}  & \xmark    & \xmark         & \xmark       & \cmark \\
EIR~\cite{ohs2025dirty}                         & \cmark    & \halfcheckmark & \xmark       & \xmark  \\
NE~\cite{aili2025development}                   & \cmark    & \xmark         & \xmark       & \xmark  \\
\textbf{LLMSpace}                               & \cmark    & \cmark         & \cmark       & \cmark  \\ \bottomrule
\end{tabular}
\end{center}
\end{wraptable} 
The large-scale deployment of LLMs has resulted in significant environmental impacts~\cite{jegham2025hungry}. Prior studies have quantified the carbon footprint of LLM training~\cite{llmcarbon} and inference~\cite{fu2025llmco2,jegham2025hungry} in terrestrial data centers. To address rising inference demand, major cloud providers have proposed solar-powered, AI-enabled LEO satellites equipped with essential peripherals and radiation-hardened AI accelerators, deployed via launch vehicles. Although solar energy reduces in-orbit operational emissions, embodied carbon from launch, satellite manufacturing, and radiation-hardened hardware fabrication may partially or fully offset these gains. 

As summarized in Table~\ref{t:related_work_comparison}, existing space-based carbon models~\cite{ohs2025dirty,aili2025development} have several limitations. First, some omit embodied emissions from critical peripherals (e.g., radiative cooling panels~\cite{ohs2025dirty}) or aggregate peripheral components into a single estimate without detailed breakdowns~\cite{aili2025development}, typically considering only one peripheral configuration. Second, prior models assume commercial COTS CMOS computing hardware and do not account for the embodied carbon and reliability implications of radiation-hardened hardware in space environments. Third, existing frameworks target conventional workloads (e.g., image processing~\cite{Bleier:MICRO2023}) and do not capture LLM-specific execution characteristics, such as the distinction between prefill and decoding phases or workload-dependent parameters including prompt length and generated token count. To address these limitations, we propose \textit{LLMSpace}, a carbon modeling framework for AI-enabled LEO satellites that jointly captures peripheral diversity, radiation-hardened hardware design, and LLM inference-specific workload behaviors.

\vspace{-0.2in}
\begin{minipage}[t]{0.24\textwidth}
\begin{equation}
E_{\mathrm{L,c}} = I_{\mathrm{L}} \cdot m_c
\label{e:carbon_launch_reentry}
\end{equation}
\end{minipage}
\begin{minipage}[t]{0.24\textwidth}
\begin{equation}
m_{\mathrm{sol}} = a_{\mathrm{sol}} \cdot A
\label{e:carbon_solar_mass}
\end{equation}
\end{minipage}
\begin{minipage}[t]{0.24\textwidth}
\begin{equation}
E_{\mathrm{sol}} = I_{\mathrm{sol}} \cdot A
\label{e:carbon_solar_manu}
\end{equation}
\end{minipage}
\begin{minipage}[t]{0.24\textwidth}
\begin{equation}
P = \mathrm{ISI} \cdot A \cdot \eta
\label{e:carbon_solar_eff}
\end{equation}
\end{minipage}
\begin{minipage}[t]{0.3\textwidth}
\begin{equation}
m_{\mathrm{bat}} = a_{\mathrm{bat}} \cdot \mathrm{cap}
\label{e:battery_mass_all}
\end{equation}
\end{minipage}
\begin{minipage}[t]{0.29\textwidth}
\begin{equation}
E_{\mathrm{bat}} = I_{\mathrm{bat}} \cdot \mathrm{cap}
\label{e:carbon_embodied_battery}
\end{equation}
\end{minipage}
\begin{minipage}[t]{0.4\textwidth}
\begin{equation}
Q = \epsilon \cdot \sigma \cdot A \cdot (T^4 - T_{\mathrm{bg}}^4)
\label{e:stefan_boltzmann_eq}
\end{equation}
\end{minipage}

\vspace{-0.1in}
\section{LLMSpace}
\label{s:llmspace}

\vspace{-0.1in}
\subsection{Embodied Carbon Modeling}
\vspace{-0.1in}

We model the embodied carbon of AI-enabled LEO satellites by accounting for emissions from launch, critical peripherals supporting computation and communication, and radiation-hardened hardware required for space-based LLM inference.

\textbf{Launch}. The embodied carbon of a payload $c$  from launch, denoted $E_{\mathrm{L,c}}$, is defined in Equation~\ref{e:carbon_launch_reentry}~\cite{aili2025development}. Let $m_c$ denote the mass of $c$, and $I_{\mathrm{L}}$ the carbon intensity per unit payload mass. For example, the Falcon-9 launch vehicle produces $\sim3.3\times10^5~\mathrm{kgCO_2e}$ per launch and supports a payload capacity of 22{,}800~kg~\cite{barker2024global}. This corresponds to a launch carbon intensity of $I_{\mathrm{L}} = 14.5~\mathrm{kgCO_2e/kg}$.

\textbf{Critical Peripherals}. AI-enabled LEO satellites span configurations from CubeSats~\cite{liddle2020space} to larger platforms~\cite{chaudhry2021laser}. These satellite classes differ in lifetime, power demand, and thermal constraints, resulting in distinct peripheral designs and embodied carbon footprints. We therefore model the embodied carbon of key peripherals, including solar arrays, batteries, and radiative cooling panels, across different satellite classes.
\begin{itemize}[leftmargin=*, nosep, topsep=0pt, partopsep=0pt]
\item \textbf{Solar Arrays.} A LEO satellite completes one orbit roughly every 90--100 minutes~\cite{prol2022position}, resulting in about $N_{\mathrm{cyc}} = 15$--16 cycles per day. Each cycle consists of 55--65 minutes of sunlight exposure for power generation and battery charging, followed by 30--35 minutes of eclipse during which stored energy is used to support LLM inference. The solar array mass is computed using Equation~\ref{e:carbon_solar_mass}, where $A$ is the total panel area, and $a_{\mathrm{sol}}$ means the areal mass density.  Key parameters, including areal mass density for Si panels, GaAs arrays, and multi-junction arrays, are summarized in Table~\ref{t:areal_density_solar}. The manufacturing embodied carbon of the solar array is computed using Equation~\ref{e:carbon_solar_manu}, where $I_{\mathrm{sol}}$ is the embodied carbon intensity.  The power $P$ generated by the array is calculated using Equation~\ref{e:carbon_solar_eff}, where $\mathrm{ISI}$ represents the incident solar irradiance at Earth orbit ($1367~\mathrm{W/m^2}$) and $\eta$ denotes the conversion efficiency. Solar array technologies differ in embodied carbon intensity, and conversion efficiency. For instance, Si panels~\cite{louwen2016cost} exhibit lower $\eta$ compared to advanced technologies, yet they provide the smallest density $a_{\mathrm{sol}}$ and the lowest manufacturing embodied carbon intensity $I_{\mathrm{sol}}$ among the considered solar array options.

\begin{table}[t!]
\centering
\begin{minipage}{0.45\linewidth}
\caption{Solar array technologies~\cite{louwen2016cost,chanoine2015integrating} (the unit of $a_{\mathrm{sol}}$ is $\mathrm{kg/m^2}$, and the unit of $I_{\mathrm{sol}}$ is $\mathrm{kgCO_2e/m^2}$).} 
\label{t:areal_density_solar}
\vspace{0.1in}
\setlength{\tabcolsep}{3pt}
\footnotesize 
\centering
\begin{tabular}{cccc}\toprule
\multirow{2}{*}{parameter}  & Si       & GaAs     & multi-   \\ 
														& panel    & array    & junction \\\midrule
$a_{\mathrm{sol}}$          & 1        & 1.1      & 1.2        \\ 
$\eta$                      & 0.17     & 0.3      & 0.5      \\
$I_{\mathrm{sol}}$          & 80       & 130      & 180      \\
\bottomrule
\end{tabular}
\end{minipage}
\hspace{0.2in}
\begin{minipage}{0.42\linewidth}
\caption{Battery technologies~\cite{liddle2020space,knap2020review,hasan2025advancing} (the unit of $\mathrm{cap}$ is kWh; the unit of $a_{\mathrm{bat}}$ is $\mathrm{kg/kWh}$; the unit of $I_{\mathrm{bat}}$ is $\mathrm{kgCO_2e/kWh}$).} 
\label{t:battery_tech_comp}
\vspace{0.1in}
\setlength{\tabcolsep}{2pt}
\footnotesize 
\centering
\begin{tabular}{cccc}\toprule
parameter                  & LFP     & NMC     & rad-hard  \\ \midrule
$\mathrm{cap}$             & 0.4     & 10      & 10        \\ 
$a_{\mathrm{bat}}$         & 8.3     & 4.5     & 5.5       \\
$I_{\mathrm{bat}}$         & 54      & 80      & 100       \\
\bottomrule
\end{tabular}
\end{minipage}
\vspace{-0.2in}
\end{table}


\item \textbf{Batteries.} Batteries provide power during eclipse periods and are subject to frequent charge--discharge cycles. Small satellites~\cite{liddle2020space} typically employ lithium iron phosphate (LFP) batteries with capacities on the order of tens to hundreds of watt-hours, whereas high-end satellites~\cite{Bleier:MICRO2023,amazon_leo_website_2025} deploy kilowatt-hour-scale nickel--manganese--cobalt (NMC) batteries with enhanced thermal management and radiation tolerance~\cite{hasan2025advancing}. The battery mass is computed using Equation~\ref{e:battery_mass_all}, where $a_{\mathrm{bat}}$ denotes the specific mass of the battery ($\mathrm{kg/kWh}$) and $\mathrm{cap}$ is the battery capacity. The manufacturing embodied carbon of the battery is calculated using Equation~\ref{e:carbon_embodied_battery}, where $I_{\mathrm{bat}}$ represents the embodied carbon intensity. As summarized in Table~\ref{t:battery_tech_comp}, different battery technologies require distinct values of $\mathrm{cap}$, $a_{\mathrm{bat}}$, and $I_{\mathrm{bat}}$. As satellite scale increases, battery embodied carbon grows substantially due to extended mission lifetimes, and increased radiation-hardening demands.

\item \textbf{Radiative Cooling Panels.} In the vacuum environment of space, heat dissipation relies exclusively on thermal radiation. The required radiator area satisfies the Stefan--Boltzmann relation~\cite{gilmore2002space}, as expressed in Equation~\ref{e:stefan_boltzmann_eq}, where $Q$ denotes the heat load, $\epsilon$ is the surface emissivity, $\sigma$ means the Stefan--Boltzmann constant, $A$ is the effective radiator area, and $T$ and $T_{\mathrm{bg}}$ represent the radiator and background temperatures, respectively. Small CubeSats typically employ passive, body-mounted radiators for sub-kilowatt thermal loads~\cite{liddle2020space}, whereas larger platforms deploy large honeycomb or heat-pipe radiators capable of dissipating multi-kilowatt heat loads~\cite{Bleier:MICRO2023,amazon_leo_website_2025}. The radiator mass ($m_{\mathrm{rad}}$) is computed using Equation~\ref{e:carbon_rad_mass}, where $a_{\mathrm{rad}}$ denotes the areal density. The manufacturing embodied carbon of radiative cooling panels is calculated using Equation~\ref{e:carbon_rad_manu}, where $I_{\mathrm{rad}}$ represents the embodied carbon intensity per unit mass. As summarized in Table~\ref{t:areal_density_raditive}, different radiative cooling technologies exhibit distinct values of $a_{\mathrm{rad}}$, $\epsilon$, and $I_{\mathrm{rad}}$. As onboard compute density increases, radiative cooling becomes a major contributor to embodied carbon, particularly for high-performance AI-enabled LEO platforms.
\end{itemize}

\begin{table}[t!]
\centering
\begin{minipage}{0.45\linewidth}
\caption{Radiative cooling panel technologies~\cite{amazon_leo_website_2025,gilmore2002space} (the unit of $a_{\mathrm{rad}}$ is $\mathrm{kg/m^2}$; the unit of $I_{\mathrm{rad}}$ is $\mathrm{kgCO_2e/kg}$).} 
\label{t:areal_density_raditive}
\vspace{0.05in}
\setlength{\tabcolsep}{2pt}
\footnotesize 
\centering
\begin{tabular}{cccc}\toprule
\multirow{2}{*}{parameter}  & passive       & aluminum     & advanced    \\ 
														& bdy-mnted     & honeycmb     & heat-pipe     \\\midrule
$a_{\mathrm{rad}}$          & 2.7           & 2.8          & 3        \\ 
$\epsilon$                  & 0.8           & 0.95         & 0.98     \\
$I_{\mathrm{rad}}$          & 12            & 13.8         & 17        \\
\bottomrule
\end{tabular}
\end{minipage}
\hspace{0.05in}
\begin{minipage}{0.45\linewidth}
\caption{CPA values of commercial COTS \& radiation-hardened process technologies (The unit is $\mathrm{kgCO_2/cm^2}$).} 
\label{t:cpa_hard_carbon}
\vspace{0.1in}
\setlength{\tabcolsep}{3pt}
\footnotesize 
\centering
\begin{tabular}{ccccc} \toprule
process (nm) & 28   & 14    & 7    & 4 \\ \midrule
COTS         & 0.9  & 1.2   & 2.15 & 3 \\
rad-hard     & 1.8  & 2.4   & -    & -  \\\bottomrule
\end{tabular}
\end{minipage}
\vspace{-0.25in}
\end{table}

\vspace{-0.25in}
\begin{minipage}[t]{\textwidth}
\centering
\begin{minipage}[t]{0.26\textwidth}
\begin{equation}
m_{\mathrm{rad}} = a_{\mathrm{rad}} \cdot A
\label{e:carbon_rad_mass}
\end{equation}
\end{minipage}
\begin{minipage}[t]{0.3\textwidth}
\begin{equation}
E_{\mathrm{rad}} = I_{\mathrm{rad}} \cdot m_{\mathrm{rad}}
\label{e:carbon_rad_manu}
\end{equation}
\end{minipage}
\begin{minipage}[t]{0.3\textwidth}
\begin{equation}
E_{\mathrm{l/m}} = \mathrm{CPA} \cdot A
\label{e:carbon_CPU_all}
\end{equation}
\end{minipage}
\end{minipage}

\textbf{Computing and Communication Hardware.} We model the embodied carbon of all hardware components required for LEO-based LLM inference, including processors, accelerators, memory, storage, and communication subsystems.
\begin{itemize}[leftmargin=*, nosep, topsep=0pt, partopsep=0pt]
\item \textbf{CPU and GPU.} The manufacturing embodied carbon of processor logic is computed using Equation~\ref{e:carbon_CPU_all}, where $A$ denotes chip area and $\mathrm{CPA}$ represents the carbon intensity per unit area~\cite{llmcarbon}. Small satellites (e.g., CubeSats) typically employ commercial COTS CPUs and GPUs with operational lifetimes of $\sim2$ years in LEO~\cite{tuxera_cubesats_2023,el2025fault,slater2020total}. For extended missions ($\sim10$ years), radiation-hardened designs adopt edgeless CMOS layouts, FB-SOI processes, and additional packaging and qualification procedures~\cite{moore_rad_2019,Jones:IEDM2023}, increasing fabrication complexity. Due to limited public data, we conservatively assume that radiation hardening doubles $\mathrm{CPA}$ at a given technology node and is available only at mature nodes, as summarized in Table~\ref{t:cpa_hard_carbon}. At the architectural level, mitigation techniques for DD and SEE—such as ECCs and TMR—introduce area overheads ranging from 10\% to 300\%. Empirical data (e.g., the RAD750 at 150\,nm with 130~mm$^2$ area~\cite{wikipedia_rad750}) suggest $\sim2\times$ area overhead relative to non-hardened designs.

\item \textbf{DDR and HBM Memory.} The manufacturing embodied carbon of commercial COTS DDR and HBM DRAM is also computed using Equation~\ref{e:carbon_CPU_all}~\cite{llmcarbon}, where $\mathrm{CPA}$ denotes the carbon intensity per GB and $A$ is the memory capacity. For 18\,nm DDR4, $\mathrm{CPA} = 0.4~\mathrm{kgCO_2e/GB}$; for HBM2, $\mathrm{CPA} = 1.8~\mathrm{kgCO_2e/GB}$. Commercial COTS DRAM typically supports missions of $\sim2$ years in LEO~\cite{slater2020total}. For extended $10$-year lifetimes, MRAM-based replacements mitigate TID, DD, and SEE vulnerabilities~\cite{lakys2011hardening}. Due to additional processing steps and magnetic materials, $\mathrm{CPA}$ increases to 0.6~$\mathrm{kgCO_2e/GB}$ (DDR-compatible) and 2.3~$\mathrm{kgCO_2e/GB}$ (HBM-compatible)~\cite{bayram2016modeling}.

\item \textbf{NAND Flash SSD.} The manufacturing embodied carbon of COTS NAND flash SSDs is also computed using Equation~\ref{e:carbon_CPU_all}~\cite{llmcarbon}. For 20\,nm NAND flash, $\mathrm{CPA} = 0.018~\mathrm{kgCO_2e/GB}$~\cite{llmcarbon}. CubeSat-class systems employ COTS SSDs with redundancy and ECC for about two-year missions~\cite{tuxera_cubesats_2023}. Radiation-tolerant configurations—including SLC operation, radiation-hardened controllers, enhanced ECCs, periodic scrubbing, and shielding—extend lifetime to 10 years~\cite{mercury_rh3480}, increasing embodied carbon to $\mathrm{CPA} = 0.16~\mathrm{kgCO_2e/GB}$~\cite{Tannu:SIGENERGY2023}. 

\item \textbf{Computing Subsystems.} Besides logic and memory devices, a computing node equipped with one or more accelerators~\cite{nvidia_hgx_h100} includes electromechanical modules, discrete electronic components (e.g., capacitors, resistors, diodes, and inductors), printed circuit boards (PCBs), interconnects, and structural elements. We approximate the manufacturing embodied carbon of these auxiliary components as 10\%~\cite{nvidia_hgx_h100} of the total embodied carbon of the computing system. Representative system masses vary by configuration. A compact NVIDIA Jetson platform weighs $\sim2$~kg~\cite{slater2020total}. A mid-range NVIDIA system integrating one or two GPUs has a mass of $\sim30$~kg~\cite{aili2025development}. High-performance platforms, such as the NVIDIA DGX A100, H100, and B100, each weigh $\sim130$~kg.

\item \textbf{Communication Subsystems and Satellites.} As modern LEO satellites are primarily communication-oriented, we jointly model the embodied carbon and mass of the satellite platform and its communication subsystem. CubeSat platforms typically employ RF-based ground-to-satellite links (GSLs) operating in UHF/S/X bands, supporting data rates from Kbps to several hundred Mbps with transmit powers of 1--20~W~\cite{palo2014expanding}. These communication payloads generally weigh 0.5--5~kg, corresponding to approximately 5--50~$\mathrm{kgCO_2e}$ embodied carbon~\cite{aili2025development}. In contrast, larger platforms (e.g., Starlink V2) adopt laser inter-satellite links (ISLs) capable of up to 200~Gbps at approximately 1~W optical power~\cite{chaudhry2021laser}. Although energy efficient during operation, laser terminals incur substantial embodied carbon (150~$\mathrm{kgCO_2e}$) and add about 28~kg per terminal to payload mass~\cite{gregory2024status}. High-capacity GSL systems further target downlink/uplink rates exceeding 96/32~Gbps~\cite{de2025satellite}, with transmit equivalent isotropically radiated power reaching 1.75~kW~\cite{aili2025development}. Under such configurations, the combined mass and manufacturing embodied carbon of the satellite and communication subsystem can be substantial; for example, a Starlink satellite weighs about 191~kg and corresponds to approximately 2940~$\mathrm{kgCO_2e}$ embodied carbon~\cite{aili2025development}.
\end{itemize}

\vspace{-0.15in}
\begin{minipage}[t]{0.45\textwidth}
\begin{equation}
O_{\mathrm{solar}} = P \cdot N_{\mathrm{cyc}} \cdot T_{\mathrm{lit}}
\label{e:carbon_solar_energy}
\end{equation}
\end{minipage}
\begin{minipage}[t]{0.45\textwidth}
\begin{equation}
O_{\mathrm{bat}} = N_{\mathrm{cyc}} \cdot \mathrm{cap} \cdot \mathrm{DoD}
\label{e:carbon_battery_operational}
\end{equation}
\end{minipage}

\vspace{-0.1in}
\subsection{Operational Carbon Modeling}
\vspace{-0.1in}

The daily energy generated by the solar array is computed according to Equation~\ref{e:carbon_solar_energy}, and the daily energy processed by the battery is given by Equation~\ref{e:carbon_battery_operational}, where $\mathrm{cap}$ denotes the battery capacity, and $\mathrm{DoD}$ indicates the depth of discharge. The energy generated by solar arrays $O_{\mathrm{solar}}$ must satisfy two constraints: (i) meeting the power demand of computing and communication subsystems and (ii) recharging the battery during sunlit periods ($T_{\mathrm{lit}}$). Conversely, the battery capacity $\mathrm{cap}$ must be sufficient to sustain subsystem operation throughout eclipse intervals.

\vspace{-0.1in}
\section{Validation}
\vspace{-0.1in}
\label{s:val}

We validate \textit{LLMSpace} by comparing its estimates against a reference model constructed from component-level embodied carbon data reported in prior work. Due to the absence of publicly available lifecycle measurements for AI-enabled LEO satellites, this reference serves as a high-fidelity bottom-up estimate rather than empirical ground truth. The evaluation thus constitutes a comparative analytical validation, assessing the completeness and consistency of \textit{LLMSpace} relative to prior frameworks. Such bottom-up aggregation is standard~\cite{pryshlakivsky2013fifteen} in lifecycle assessment when system-level measurements are unavailable. We estimate the embodied carbon of AI-enabled Starlink-V1 satellites equipped with an NVIDIA DGX H100 system comprising eight GPUs under multiple configurations and lifetimes, while the operational energy model is validated in~\cite{ozcan2025quantifying}. Launch emissions of Falcon-9 are adopted from~\cite{Wikipedia:Falcon9}. The manufacturing embodied carbon and mass of the DGX H100 system, excluding thermal components, are obtained from~\cite{nvidia_hgx_h100}. Embodied carbon and mass parameters for solar arrays, batteries, and radiative cooling panels are derived from~\cite{louwen2016cost,chanoine2015integrating,liddle2020space,knap2020review,hasan2025advancing,amazon_leo_website_2025,gilmore2002space}. Commercial COTS DGX H100 chips are fabricated at 4\,nm, whereas radiation-tolerant counterparts are assumed to use a 14\,nm process~\cite{boss14nm}. The CPA of radiation-hardened logic is estimated using~\cite{moore_rad_2019}. To capture the combined impact of node scaling and architectural redundancy (e.g., ECC and TMR), we conservatively assume an overall $\sim15\times$ chip area increase relative to COTS, with sensitivity analyzed in Section~\ref{s:use}. The carbon intensity per unit capacity of radiation-hardened MRAM and SSD NAND flash is adopted from~\cite{bayram2016modeling,mercury_rh3480}. The manufacturing embodied carbon of the Starlink-V1 satellite and its communication subsystem is aggregated following~\cite{aili2025development}. Radiation-tolerant GPUs operate at higher supply voltages and incur additional capacitance and redundancy overheads, increasing operational energy. To conservatively model this effect, we assume a $\sim2\times$ power overhead relative to COTS devices, with sensitivity analyzed in Section~\ref{s:use}. LLM inference energy is estimated using the random-forest-based \textit{Vidur-Energy} model~\cite{ozcan2025quantifying}.

\newcolumntype{Y}{>{\centering\arraybackslash}X}
\begin{table}[t!]
\caption{The embodied carbon validation of Starlink-V1 LEO satellites equipped with a DGX H100 node and launched by Falcon-9 (unit: $\mathrm{tCO_2e}$).} 
\label{t:cpa_valid_carbon}
\vspace{0.1in}
\setlength{\tabcolsep}{2pt}
\footnotesize 
\centering
\begin{tabularx}{0.8\textwidth}{c*{8}{Y}} \toprule
                   & \multicolumn{6}{c}{\textbf{COTS}} & \multicolumn{2}{c}{\textbf{rad-hard}}\\ \cmidrule(lr){2-7}\cmidrule(lr){8-9}
\textbf{component} & \multicolumn{2}{c}{\textbf{EIR}~\cite{ohs2025dirty}} & \multicolumn{2}{c}{\textbf{NE}~\cite{aili2025development}} & \multicolumn{2}{c}{\textbf{LLMSpace}} & \multicolumn{2}{c}{\textbf{LLMSpace}} \\\cmidrule(lr){2-3}\cmidrule(lr){4-5}\cmidrule(lr){6-7}\cmidrule(lr){8-9}
                   & manu & launch  & manu & launch  & manu & launch  & manu & launch \\\midrule
solar array        & 3.25 & 0.59    & -    & -       & 3.25 & 0.59    & 6.5  & 1.18   \\\midrule
battery            & 0.63 & 0.51    & -    & -       & 0.63 & 0.51    & 1.26 & 1.03   \\\midrule
cooling panel      & -    & -       & 2.29 & 2.41    & 2.29 & 2.41    & 4.58 & 4.81   \\\midrule
computing HW       & 0.96 & 1.89    & 0.96 & 1.89    & 0.96 & 1.89    & 5.16 & 1.89   \\\midrule
net+satellite      & 1.63 & 2.41    & 2.94 & 2.41    & 1.63 & 2.41    & 1.63 & 2.41   \\\midrule
\textbf{total}    & \multicolumn{2}{c}{11.87} & \multicolumn{2}{c}{12.9} & \multicolumn{2}{c}{16.57} & \multicolumn{2}{c}{30.45} \\ \midrule
\multicolumn{9}{c}{constructed embodied carbon references for Starlink-V1 configs are 18.3 (COTS) and 32.5 (rad-hard)}\\ \midrule
prediction $\mathbf{\Delta}$ &  \multicolumn{2}{c}{-35.1\%} &  \multicolumn{2}{c}{-29.5\%} &  \multicolumn{2}{c}{\textbf{-9.4\%}} &  \multicolumn{2}{c}{\textbf{-6.3\%}}\\
\bottomrule
\end{tabularx}
\vspace{-0.2in}
\end{table}

We evaluate two configurations: commercial COTS and radiation-hardened (rad-hard). For comparison with prior tools, EIR~\cite{ohs2025dirty} and NE~\cite{aili2025development}, we adopt a Starlink V1 baseline integrating a DGX H100 node with a Si solar array, NMC battery, and aluminum honeycomb radiative cooling panel, assuming a 2-year lifetime. To ensure a fair comparison, identical embodied carbon parameters for peripheral components are used across all methods. The COTS configuration assumes a 10~kW computing power budget based on reported DGX H100 inference workloads~\cite{newkirk2025empirically}. The rad-hard configuration employs a 14\,nm radiation-tolerant design with MRAM and Flash SSD, and assumes a 20~kW power budget to account for increased voltage, capacitance, and architectural overheads. This configuration targets a 10-year operational lifetime, requiring correspondingly larger supporting peripherals.

As shown in Table~\ref{t:cpa_valid_carbon}, we estimate the manufacturing (manu) and launch embodied carbon of key peripherals, with battery capacity sized at 80\% DoD. EIR~\cite{ohs2025dirty} omits radiative cooling panels, resulting in a $-35.1\%$ deviation from the reference. NE~\cite{aili2025development} aggregates solar array and cooling panel emissions, failing to capture the increased peripheral requirements of the DGX H100 node and yielding a $-29.5\%$ deviation. By explicitly modeling all critical peripherals, \textit{LLMSpace} achieves the closest estimate, with a $-9.4\%$ deviation primarily due to underestimation of launch emissions. For extended lifetimes, the radiation-hardened configuration increases computing power (to 20~kW) and peripheral sizes, further raising embodied carbon. Under this configuration, \textit{LLMSpace} maintains high accuracy, with only a $-6.3\%$ deviation from the reference value.

\begin{figure}[h!]
\vspace{-0.1in}
\centering
\subcaptionbox{DGX-H100.\label{f:carbon_year_h100}}
{\includegraphics[width=1.6in]{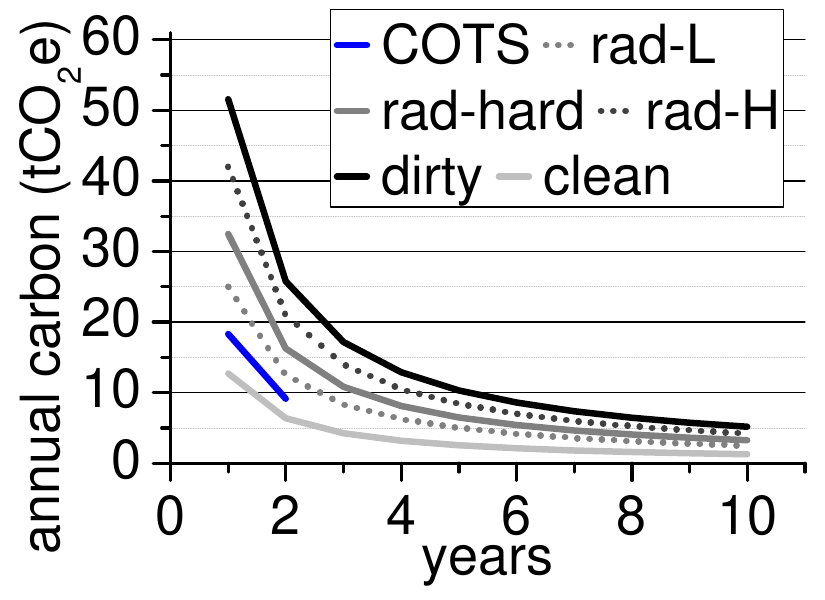}}
\hspace{0.01in}
\subcaptionbox{Carbon breakdown.\label{f:carbon_break_both}}
{\includegraphics[width=1.6in]{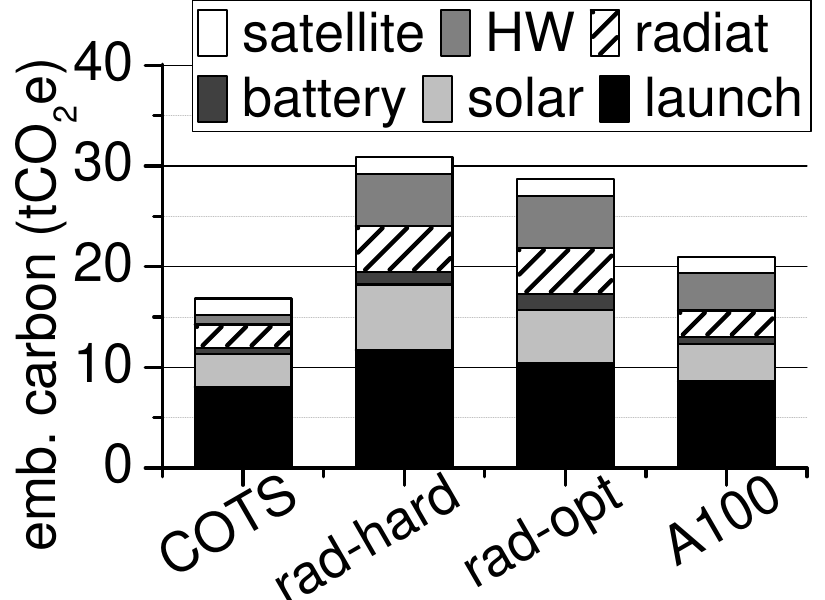}}
\hspace{0.01in}
\subcaptionbox{Jetson Nano.\label{f:carbon_year_agx}}
{\includegraphics[width=1.6in]{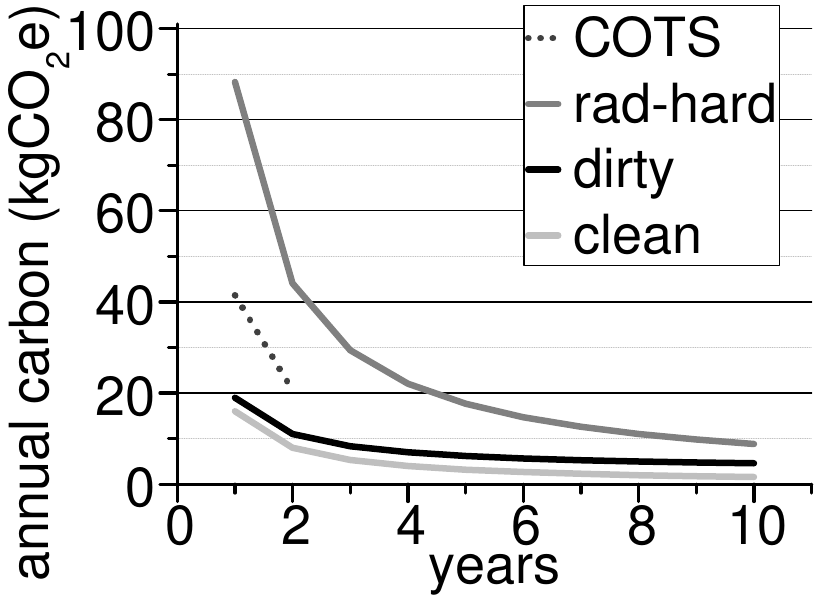}}
\vspace{-0.1in}
\caption{The carbon footprint comparison between orbital and terrestrial data centers.}
\label{f:carbon_space_terrestrial}
\vspace{-0.1in}
\end{figure}

\vspace{-0.1in}
\section{Use Case Studies}
\vspace{-0.1in}
\label{s:use}

We use \textit{LLMSpace} to demonstrate the following case studies.

\vspace{-0.1in}
\subsection{Carbon footprint comparison between orbital and terrestrial data centers}
\vspace{-0.1in}

\textbf{Orbital vs.\ Terrestrial Carbon}. Figure~\ref{f:carbon_space_terrestrial} compares the carbon footprint of orbital and terrestrial data centers. As shown in Figure~\ref{f:carbon_year_h100}, we report the annualized emissions of a COTS DGX H100 in terrestrial data centers powered by grids with different carbon intensities (\textit{clean}: 20~$\mathrm{gCO_2e/kWh}$; \textit{dirty}: 380~$\mathrm{gCO_2e/kWh}$), and a DGX H100 deployed on a Starlink-V1 satellite under COTS (\textit{COTS}) and radiation-hardened (\textit{rad-hard}) configurations. Terrestrial embodied carbon follows~\cite{aili2025development}. To assess sensitivity to hardening overheads, we consider a lower bound (\textit{rad-L}: $10\times$ chip area, $1.5\times$ power) and an upper bound (\textit{rad-H}: $20\times$ chip area, $3\times$ power). Annualized emissions decrease with lifetime due to amortization; however, \textit{COTS} ceases operation after $\sim2$ years in orbit due to TID, DD, and SEE. The \textit{rad-hard}, \textit{rad-L}, and \textit{rad-H} configurations lie between the \textit{clean} and \textit{dirty} terrestrial cases, indicating that radiation-hardened AI-enabled LEO satellites can partially alleviate the energy and carbon burden of large-scale LLM inference on Earth.


\textbf{Peripheral Carbon Dominance}. Figure~\ref{f:carbon_break_both} presents the embodied carbon breakdown of \textit{COTS} and \textit{rad-hard} configurations. Compared with \textit{COTS}, radiation-hardened DGX H100 nodes incur higher hardware embodied carbon due to fabrication and resilience mechanisms (see ``HW''). More importantly, their elevated power demand necessitates larger solar arrays, batteries, and radiative cooling panels, which dominate both manufacturing and launch-related embodied emissions. We further construct an optimized configuration, \textit{rad-opt}, by adopting a multi-junction solar array and a radiation-hardened battery. Despite higher manufacturing carbon intensities, these peripherals reduce total embodied carbon by 7\% relative to \textit{rad-hard} due to lower mass and reduced launch emissions. These results highlight the importance of minimizing power demand and co-optimizing peripheral design; notably, only \textit{LLMSpace} enables systematic evaluation across multiple peripheral options.

\textbf{Small GPU Inefficiency in Orbit}. 
Compact GPU platforms (e.g., NVIDIA Jetson Nano) have been deployed in space~\cite{slater2020total}. The Jetson Nano integrates a 4-core Cortex-A57 CPU and a 128-CUDA-core GPU with 4\,GB LPDDR4 memory. Figure~\ref{f:carbon_year_agx} compares its carbon footprint in orbital and terrestrial settings. We report annualized emissions for a COTS Jetson Nano in terrestrial data centers under different grid intensities, and for orbital deployments under COTS and radiation-hardened (\textit{rad-hard}) configurations. Although annualized emissions decrease with longer lifetimes due to amortization, both orbital configurations consistently exceed terrestrial \textit{clean} and \textit{dirty} cases. This is because Jetson Nano is dominated by manufacturing embodied carbon, with operational emissions contributing only 10--20\%~\cite{fu2025co2}. While space deployment removes operational grid emissions, it significantly increases embodied carbon due to launch and hardening overheads. Thus, deploying small GPU platforms terrestrially is more carbon-efficient.

\vspace{-0.1in}
\subsection{LLM-specific workload analysis}
\vspace{-0.1in}

\begin{figure}[t!]
\centering
\subcaptionbox{Prompt Length.\label{f:input_token}}
{\includegraphics[width=1.6in]{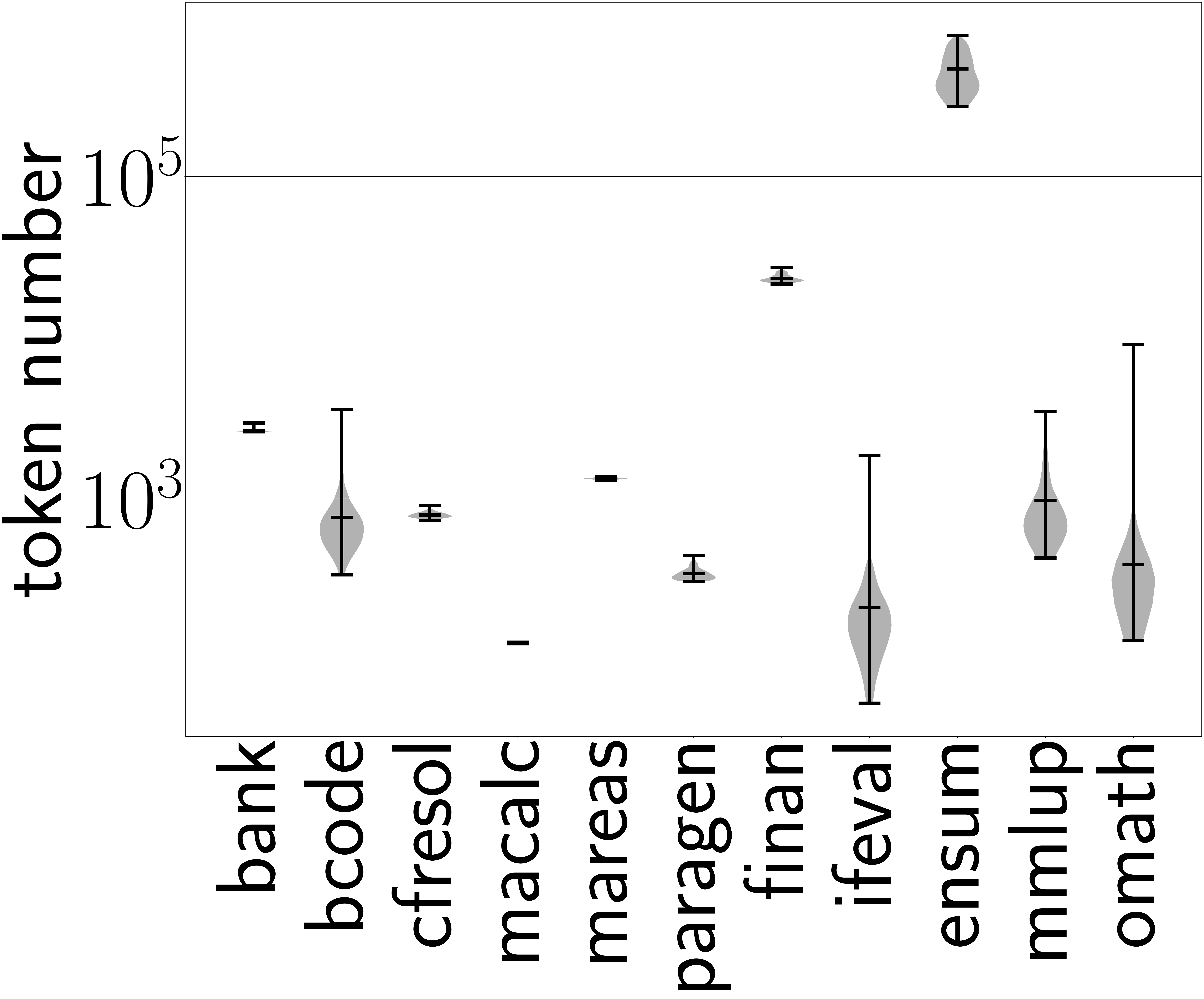}}
\hspace{0.01in}
\subcaptionbox{Generated Token \#.\label{f:gen_token}}
{\includegraphics[width=1.6in]{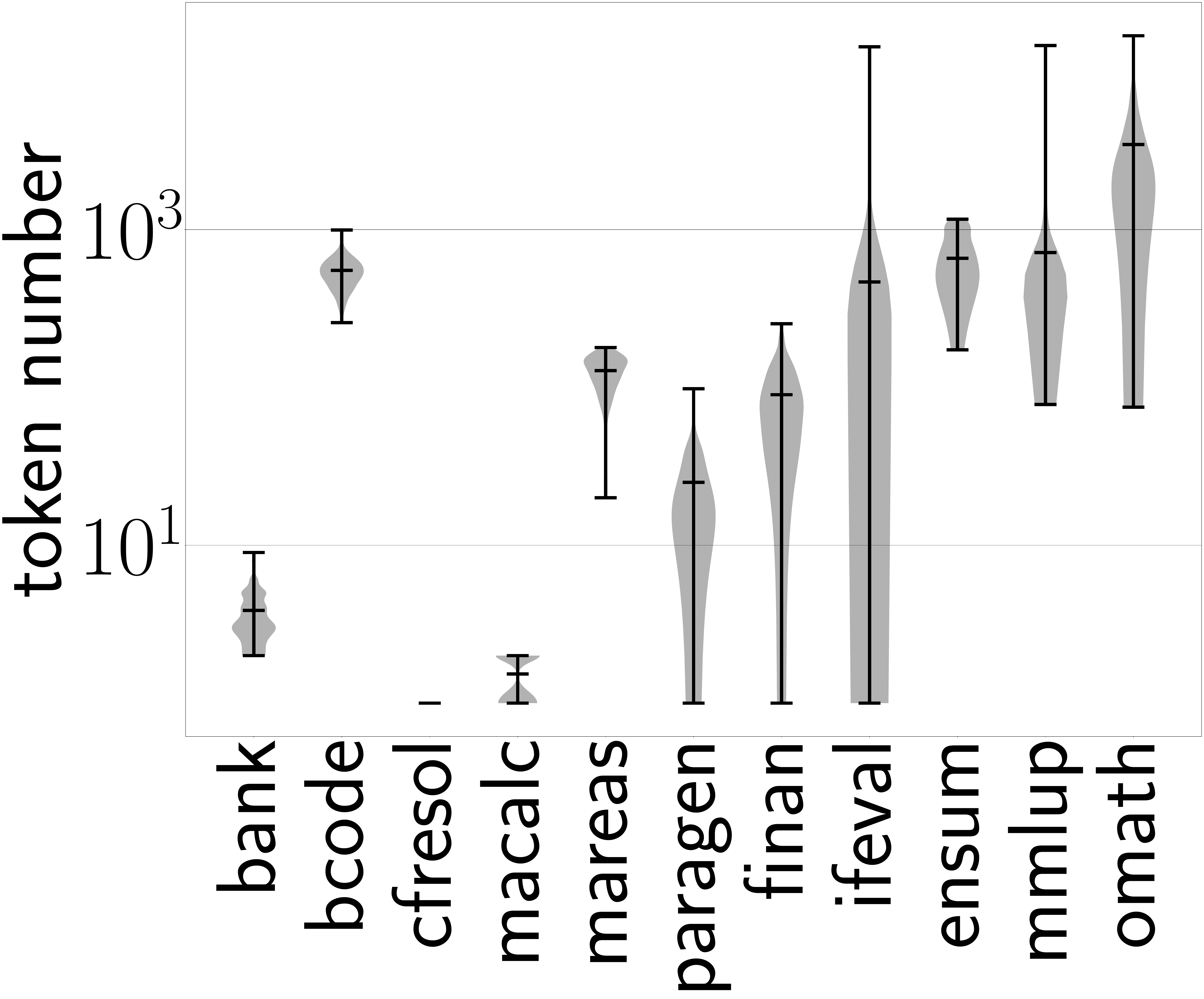}}
\hspace{0.01in}
\subcaptionbox{Inference Energy.\label{f:infer_energy}}
{\includegraphics[width=1.65in]{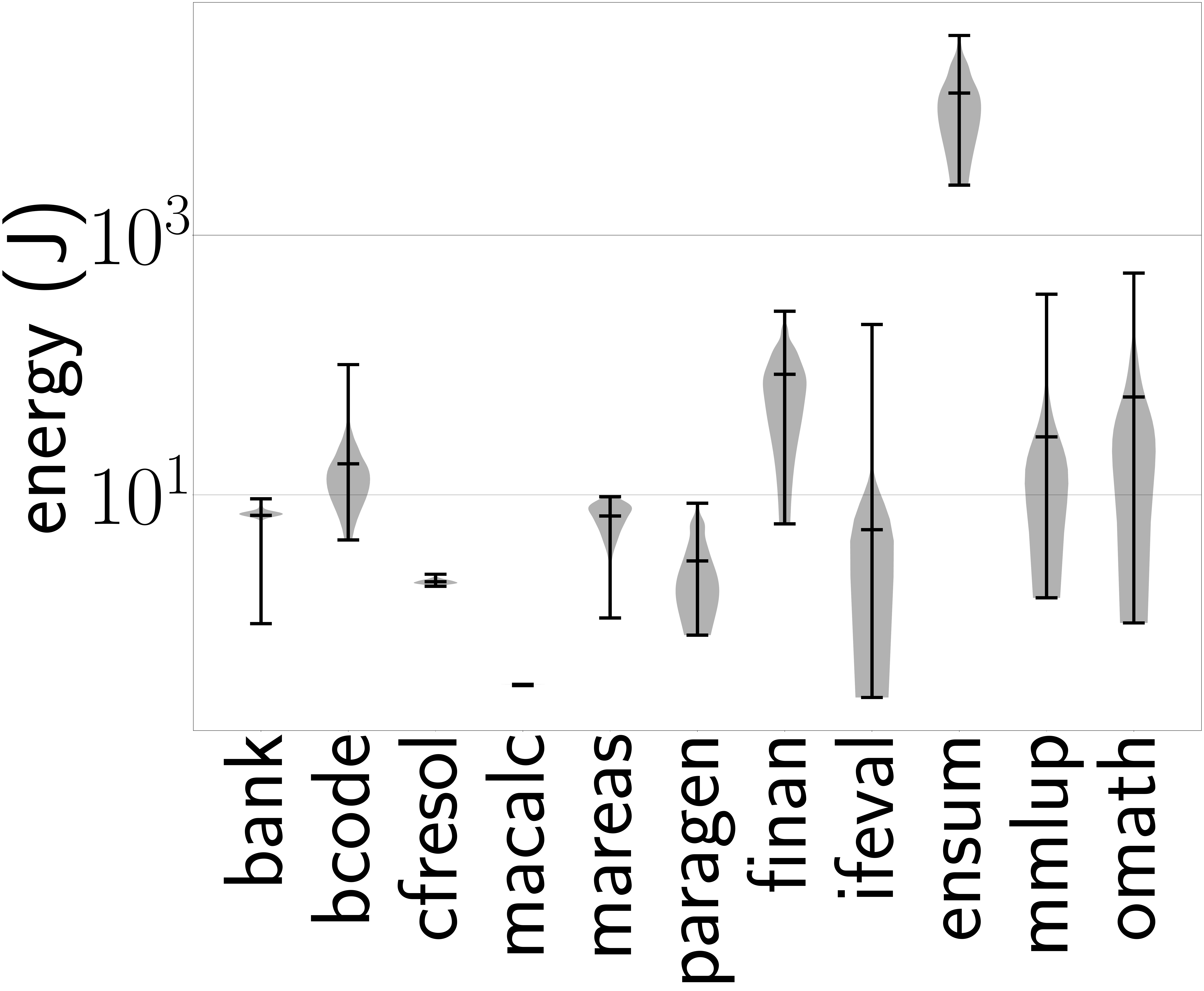}}
\vspace{-0.1in}
\caption{The operational energy consumption of LLM inferences on various benchmarks.}
\label{f:infer_energy_all}
\vspace{-0.2in}
\end{figure}

\textbf{Operational Energy}. To characterize LLM inference workloads, we adopt the HELM benchmark~\cite{liang2022holistic} and select 11 representative tasks. We select 11 representative tasks: banking77 (\textit{bank}), bigcodebench (\textit{bcode}), cleva:coreference\_resolution (\textit{cfresol}), cleva:mathematical\_calculation (\textit{macalc}), cleva:mathematical\_reasoning (\textit{mareas}), cleva:paraphrase\_generation (\textit{paragen}), financebench (\textit{finan}), ifeval (\textit{ifeval}), infinite\_bench\_en\_sum (\textit{ensum}), mmlu\_pro (\textit{mmlup}), and omni\_math (\textit{omath}).
\begin{wrapfigure}[14]{r}{0.38\linewidth}
\vspace{-0.1in}
\centering
\includegraphics[width=1.7in]{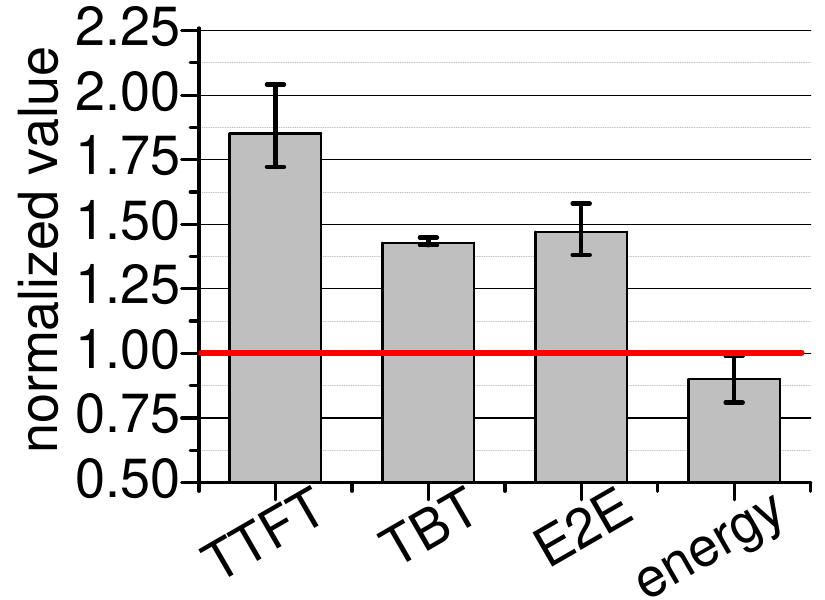}
\caption{The latency and operational energy comparison between DGX-H100 and -A100 nodes (normalized to DGX-H100).} 
\label{f:carbon_latency_sum}
\end{wrapfigure} 
The distributions of input prompt lengths and generated token counts are shown in Figure~\ref{f:input_token} and Figure~\ref{f:gen_token}. Error bars denote min–max across requests. We run inference using the \textit{codellama/CodeLlama-34b-Instruct-hf} model with bfloat16 precision on a single GPU of an NVIDIA DGX H100 node (batch size = 1). The resulting energy consumption is reported in Figure~\ref{f:infer_energy}. For a fixed number of output tokens, longer prompts increase energy consumption. However, total energy is dominated by the decoding phase~\cite{wang2025systematic}, making generated token count the primary driver of energy. The energy required for transmitting requests and responses via terrestrial equipment is $\sim0.5~\mu\mathrm{J/bit}$~\cite{ohs2025dirty}, which is orders of magnitude smaller than inference energy across all tasks. These results indicate that, despite non-negligible transmission overhead, offloading LLM inference with large output sizes to AI-enabled LEO satellites can reduce terrestrial energy demand and associated carbon emissions.


\textbf{Latency–Carbon Tradeoff}. As shown in Figure~\ref{f:carbon_break_both}, we build a configuration denoted as \textit{A100} by replacing the radiation-hardened DGX-H100 node in the \textit{rad-hard} configuration with a radiation-hardened DGX-A100 node. We run the same experiments on an A100 GPU, and estimate the energy results using \textit{Vidur-Energy}~\cite{ozcan2025quantifying}. Compared with \textit{rad-hard}, \textit{A100} reduces the total embodied carbon footprint by $\sim32\%$. This reduction is mainly attributed to the lower inference power consumption of DGX-A100 (about 11~kW)~\cite{kakolyris2024slo}, which enables smaller and lighter solar arrays, batteries, and radiative cooling panels. However, the lower peak compute throughput and reduced HBM bandwidth of DGX-A100 increase both prefill and decoding latency in LLM inferences. As shown in Figure~\ref{f:carbon_latency_sum}, compared with \textit{rad-hard}, \textit{A100} increases the time to first token (TTFT), time between tokens (TBT), and end-to-end (E2E) latency by 85\%, 43\%, and 47\%, respectively, on average across all evaluated LLM tasks. Error bars indicate min–max across tasks. The TTFT increase is more pronounced for tasks with longer prompts, while TBT and E2E latency increases are greater for tasks with larger numbers of generated tokens. Despite these latency penalties, \textit{A100} reduces inference operational energy by about 10\% on average, with larger savings for workloads generating more tokens. Thus, by trading inference latency for lower power demand, AI-enabled LEO satellites equipped with radiation-hardened DGX-A100 nodes provide a more sustainable solution for supporting LLM inference.

\vspace{-0.1in}
\section{Conclusion}
\label{s:conclusion}
\vspace{-0.1in}

This paper presents \textit{LLMSpace}, a carbon footprint modeling framework for LLM inference on AI-enabled LEO satellites. \textit{LLMSpace} jointly models embodied and operational emissions, radiation-tolerant hardware effects, and LLM-specific workload characteristics. Our analysis shows that launch and peripheral manufacturing dominate embodied carbon, while increased power demand from radiation-tolerant designs significantly enlarges peripheral subsystems and their associated emissions. Case studies indicate that space-based LLM inference can reduce reliance on terrestrial electricity for large-scale workloads, although its benefits depend critically on hardware efficiency and system configuration. We highlight the following considerations:

\begin{itemize}[leftmargin=*, nosep, topsep=0pt, partopsep=0pt]

\item \textbf{Limitations.}  
\textit{LLMSpace} relies on several assumptions due to limited publicly available data. In particular, radiation-tolerant hardware overheads are approximated using nominal scaling factors for chip area and power, which may not fully capture advanced implementations. Likewise, launch emissions and component-level embodied carbon are derived from literature-based estimates rather than direct measurements. Consequently, the validation is comparative and analytical, rather than empirical, and these assumptions define the current scope of applicability.

\item \textbf{Future Work.}  
Future work can enhance both fidelity and scope. First, incorporating empirical data on advanced-node radiation-tolerant hardware would refine area, power, and carbon estimates. Second, extending the framework to multi-satellite constellations and network-level energy modeling would enable system-wide analysis. Third, supporting additional workloads, such as in-orbit LLM fine-tuning or distributed inference, would broaden applicability. Finally, integrating real-world measurements as they become available will strengthen empirical validation.

\end{itemize}

Overall, \textit{LLMSpace} provides a systematic foundation for evaluating the sustainability of orbital AI systems and reveals key trade-offs among hardware design, operational efficiency, and carbon footprint impact.

\bibliography{leo}
\bibliographystyle{unsrt}

\end{document}